\documentclass[10pt,twocolumn,letterpaper]{article}

\usepackage[pagenumbers]{cvpr} % To force page numbers, e.g. for an arXiv version

\usepackage{graphicx}
\usepackage{amsmath}
\usepackage{amssymb}
\usepackage{booktabs}
\usepackage{pifont}

\usepackage[pagebackref,breaklinks,colorlinks]{hyperref}

\usepackage{multirow}
\usepackage{booktabs}
\usepackage[capitalize]{cleveref}
\crefname{section}{Sec.}{Secs.}
\Crefname{section}{Section}{Sections}
\Crefname{table}{Table}{Tables}
\crefname{table}{Tab.}{Tabs.}

\def\cvprPaperID{51} % *** Enter the CVPR Paper ID here
\def\confName{CVPR}
\def\confYear{2023}

\begin{document}

\title{Cross-View Hierarchy Network for Stereo Image Super-Resolution}

\author{Wenbin Zou$^{1}$, Hongxia Gao$^{1, }\thanks{Corresponding author}$, Liang Chen$^2$, Yunchen Zhang$^{2}$, Mingchao Jiang$^3$, Zhongxin Yu$^2$, Ming Tan$^2$\\
South China University of Technology.$^1$ Fujian Normal University.$^2$ GAC R\&D Center.$^3$\\
{\tt\small alexzou14@foxmail.com, hxgao@scut.edu.cn, cl\_0827@126.com, jiangshaoyu1993@gmail.com,}\\
{\tt\small cydiachen@cydiachen.tech, wuyizhizi555@163.com, qsz20211396@student.fjnu.edu.cn}
}

\maketitle

%%%%%%%%% ABSTRACT
\begin{abstract}
   Stereo image super-resolution aims to improve the quality of high-resolution stereo image pairs by exploiting complementary information across views. To attain superior performance, many methods have prioritized designing complex modules to fuse similar information across views, yet overlooking the importance of intra-view information for high-resolution reconstruction. It also leads to problems of wrong texture in recovered images. To address this issue, we explore the interdependencies between various hierarchies from intra-view and propose a novel method, named \textbf{C}ross-\textbf{V}iew \textbf{H}ierarchy Network for \textbf{S}tereo Image \textbf{S}uper-\textbf{R}esolution (CVHSSR). Specifically, we design a cross-hierarchy information mining block (CHIMB) that leverages channel attention and large kernel convolution attention to extract both global and local features from the intra-view, enabling the efficient restoration of accurate texture details. Additionally, a cross-view interaction module (CVIM) is proposed to fuse similar features from different views by utilizing cross-view attention mechanisms, effectively adapting to the binocular scene. Extensive experiments demonstrate the effectiveness of our method. CVHSSR achieves the best stereo image super-resolution performance than other state-of-the-art methods while using fewer parameters. The source code and pre-trained models are available at \url{https://github.com/AlexZou14/CVHSSR}.
\end{abstract}

%%%%%%%%% BODY TEXT
\section{Introduction}
\label{sec:intro}

\begin{figure}
	\centering
	\includegraphics[width=7cm]{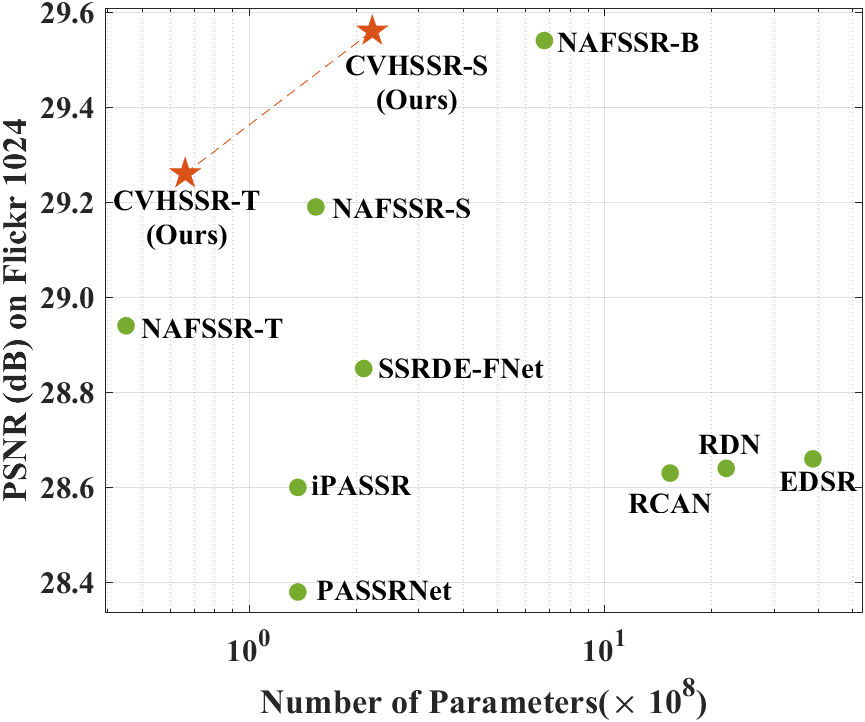}\\
        \caption{Comparision on the trade-off between model parameters and PSNR for 2× stereo SR on Flickr1024 \cite{Flickr1024} test set. Our CVHSSR model family achieves state-of-the-art performance with significantly fewer parameters, indicating its superior efficiency and effectiveness for stereo super-resolution tasks.}
	% \caption{Parameters vs. PSNR of models for 2× stereo SR on Flickr1024 \cite{Flickr1024} test set. Our CVHSSR families achieve the state-of-the-art performance with fewer parameters.}
	\label{ComparsionMultiAdd}
\end{figure}

Stereo image technology has made significant strides, leading to the successful application of stereo images in a variety of 3D scenarios, such as augmented reality (AR), virtual reality (VR), and autonomous driving. However, stereo imaging devices, such as the dual cameras on mobile phones, are subject to certain constraints that can result in the production of stereo image pairs with low resolution (LR). Stereo image super-resolution (SR), which has attracted much attention in recent years, aims to generate high-resolution stereo image pairs from their low-resolution counterparts to significantly enhance their visual perception. Therefore, this research has great potential to enhance the user experience in deploying immersive services. 

In recent years, numerous deep learning-based algorithms for stereo image SR have been proposed, following the widespread use of convolution neural network (CNN) based methods in single image super-resolution (SISR) \cite{EDSR, drfn, RCAN, SAN,liang2021swinir} tasks. Unlike SISR, which primarily focuses on finding similar textures within an image, stereo image SR must consider both intra-view and inter-view information, both of which play critical roles in stereo image reconstruction. Existing methods typically develop complex networks and loss functions to effectively fuse information from two viewpoints. For example, Jeon et al. \cite{Jeon} learned the parallax prior in stereo datasets through a two-stage network to recover high-resolution images. Wang et al. \cite{PASSR} introduced a parallax attention mechanism that incorporates global receptive fields to further improve network performance. Song et al. \cite{song} proposed a self and parallax attention mechanism to reconstruct high-quality stereo image pairs. Recently, Chu et al. \cite{nafssr} used an efficient nonlinear activation-free block and cross-view attention module, achieving the best performance and winning first place in NTIRE2022 \cite{ntire2022}. 

Although the existing stereo image SR methods have achieved impressive performance, they have not fully explored the rich hierarchy features of the intra-view, which could affect the information transfer between cross-views. Therefore, an interesting research question remains on how to effectively utilize both global and local features from stereo image pairs to further improve the quality of stereo image SR reconstruction.

In this paper, we propose a novel method to address the issue of unexplored intra-view hierarchical features in stereo image SR. The proposed method, named Cross-View Hierarchical Network for stereo image SR (CVHSSR), aims to extract rich feature representations from intra-views at different hierarchies and fuse them to enhance the performance of stereo image SR. To achieve this, we introduce two core modules: the cross-hierarchy information mining block (CHIMB) and the cross-view interaction module (CVIM), which explore and fuse similar features from different hierarchies across views. Specifically, the CHIMB module is designed to model and recover intra-view information at various hierarchies, utilizing the large kernel convolution attention and the channel attention mechanism. On the other hand, the CVIM module effectively integrates similar information from different views by utilizing the cross-view attention mechanism. By exploiting these modules, the CVHSSR can incorporate more diversified feature representations from different spatial levels of the two views, resulting in enhanced SR reconstruction quality.

The key contributions of this work are summarized as follows:
\begin{itemize}
    \item We propose the CHIMB to efficiently extract hierarchy information from intra-views. In contrast to the NAFBlock used in NAFNet \cite{nafnet}, CHIMB models global and local information from intra-view by using channel attention and large kernel convolution attention, effectively helping the network to restore the correct texture features.
    \item We have designed the CVIM to fuse similar information from different views in our method. Unlike the cross-view attention mechanism used in other methods, CVIM utilizes depth-wise convolution to capture similar information from intra-view, then facilitating cross-view information fusion.
    \item Based on CHIMB and CVIM, we propose a simple yet effective method for stereo image SR. Our approach achieves state-of-the-art performance with fewer parameters, as shown in Figure \ref{ComparsionMultiAdd}. Extensive experiments confirm the validity of our proposed CVHSSR.
\end{itemize}

\begin{figure*}[ht]
\vspace{-0.3cm}
\setlength{\abovecaptionskip}{0.1cm} 
\setlength{\belowcaptionskip}{-0.5cm}%调整caption与下文的距离
	\centering
	\includegraphics[width=\textwidth]{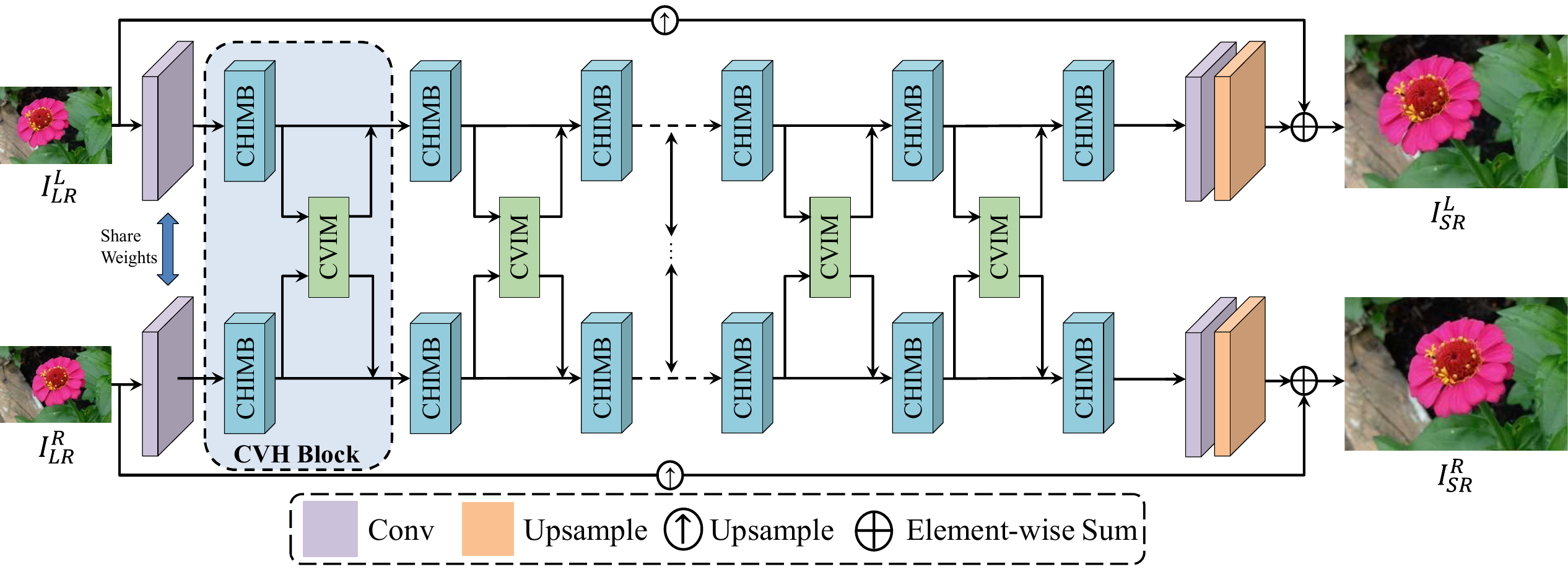}\\
	\caption{The architecture of Cross-View Hierarchy Network for Stereo Image SR (CVHSSR), which incorporates two core modules: (a) Cross-Hierarchy Information Mining Block (CHIMB), and (b) Cross-View Interaction Module (CVIM).}
	\label{network1}
\end{figure*}

%------------------------------------------------------------------------
\section{Related Work}

%-------------------------------------------------------------------------
\subsection{Single Image Super-Resolution}

Image super-resolution is a regression problem that maps a low-resolution image to its corresponding high-resolution image. Since Dong et al. \cite{SRCNN} introduce CNN into the SISR field with their pioneering work SRCNN, CNN-based methods have been proven to achieve impressive performance in SISR tasks. On this basis, Lim et al. \cite{VDSR} further improved network performance by deepening the network and increasing the dimension of intermediate features. With the  development of deep learning technology, the researchers use residual and dense connections \cite{RDN, EDSR, DRCN, drfn} to control network information flow, thereby obtaining better image reconstruction performance. However, these methods did not consider the importance of different features, leading to redundant network design. Therefore, RCAN \cite{RCAN} introduced channel attention mechanisms to model the interdependence between feature channels and adaptively rescale the features of each channel. Subsequently, various attention mechanisms were proposed to enhance network expression ability, including spatial attention \cite{DaiRBAN, HAN}, second-order attention\cite{SAN}, non-local attention\cite{CSNLN, nonlocal}, and large kernel convolution attention \cite{man}. 

Recently, the Transformer has achieved great success in the field of vision. Therefore, many researchers \cite{eformer,liang2021swinir,swinfir, lu2022transformer, uformer} introduce the Transformer into SISR tasks. With the powerful learning ability of Transformers, Transformer-based methods have achieved state-of-the-art performance in the field of single-image super-resolution. Despite the consistently improved performance, SISR cannot utilize complementary information from different views in stereo image pairs, which limits the performance of stereo image super-resolution.
%------------------------------------------------------------------------
\subsection{Stereo Image Super-Resolution}

Instead of SISR method, which only has access to context information from intro, stereo image SR can leverage the additional information provided by the cross-view information to enhance SR performance. However, the presence of binocular disparity between the left and right views in a stereo image pair can pose a significant challenge to the fusion of information across views. Thus, Jeon et al. \cite{Jeon} proposed the first deep learning-based model for stereo image SR (namely, StereoSR). This approach addresses the challenge of fusing complementary features of the left and right views by concatenating the left image and a stack of right images with predefined shifts. On this basis, Wang et al. \cite{PASSR, PASSRNet} introduced a parallax attention module (PAM) to model stereo correspondence by effectively capturing global contextual information along the epipolar line. These methods outperforms StereoSR and exhibits greater flexibility in accommodating disparity variation. In pursuit of more refined stereo correspondence, Song et al. \cite{song} extended the parallax-attention mechanism to propose SPAM, which aggregates information from both the primary and cross views to generate stereo-consistent image pairs. Yan et al. \cite{yan2020disparity} proposed a domain adaptive stereo SR network (DASSR) to achieve both stereo image SR and stereo image deblurring tasks. Xu et al. \cite{xu2021deep} introduced the concept of bilateral grid processing within a CNN framework, thereby proposing a bilateral stereo SR network. Then, Wang et al. \cite{iPASSR} enhanced PASSRNet (ie. iPASSRNet) by leveraging the symmetry cues present in stereo image pairs. Ma et al. proposed a perception-oriented StereoSR framework, which aims to restore stereo images with improved subjective quality. More recently, Chu et al. \cite{nafssr} developed the NAFSSR network by utilizing nonlinear activation-free blocks \cite{nafnet} for intra-view feature extraction and PAM for cross-view feature interaction, which achieves the champion in the NTIRE 2022 Stereo Image SR Challenge \cite{ntire2022}.

Although existing methods have achieved superior performance in stereo image SR, they typically focus on modeling cross-view information while neglecting the hierarchical similarity relationships from intra-view. To address this limitation, we propose to leverage both local and global hierarchical feature representations to further improve the performance of state-of-the-art stereo image SR methods.

\section{Cross-View Hierarchy Network}

\subsection{Overall Framwork}
\label{3.1}
%Our CVHSSR mainly consists of three components: the cross-hierarchy information mining block (CHIMB), the cross-view interaction modules (CVIM), and reconstruction.
To avoid complex network designs that require a large number of parameters and computational effort, we adopt a simple weight-sharing two-branch network structure to recover the left and right view images, as illustrated in Figure \ref{network1}. Our CVHSSR mainly consists of four components: the shallow feature extraction, the cross-hierarchy information mining block (CHIMB), the cross-view interaction modules (CVIM), and stereo image reconstruction. Specifically, the CHIMB is designed to extract similar features both locally and globally from the intra-view image, effectively restoring accurate texture details. The CVIM is mainly used to fuse features from two viewpoints. More details of CHIMB and CVIM are described in Sections \ref{3.2} and \ref{3.3}.

Firstly, given an input stereo low-resolution images $I^L_{LR}, I^R_{LR} \in \mathcal{R}^{H\times W\times 3}$, CVHSSR first applies a convolution to obtain two-views shallow feature $F^L_0, F^R_0 \in \mathcal{R}^{H\times W\times C}$, where $H \times W$ denotes the spatial dimension and $C$ is the number of channels. It can be formulated as:
\begin{equation}
    F^{L,R}_0 = H_{conv}(I^L_{LR}, I^R_{LR}),
\end{equation}
where $H_{conv}$ denotes $3\times 3$ convolution operation. 

Next, we integrate CHIMB and CVIM into a cross-view hierarchy (CVH) block, which not only extracts deep intra-view features but also fuses information from different view images. Therefore, we stack $N$ CVH blocks to obtain output features that incorporate information from multiple perspectives. It can be expressed as:
\begin{equation}
    \begin{aligned}
        F^{L,R}_{out} = H_{CVH}^N(H_{CVH}^{n-1}(\cdots H_{CVH}^1(F^{L, R}_0))\cdots), \\
        F^{L,R}_{i+1} = H_{CVH}(F^{L,R}_{i}) = H^i_{CV}(H^i_{CH}(F^{L,R}_{i})),
    \end{aligned}
\end{equation}
where $H_{CVH}, H_{CV}$, and $H_{CH}$ denote CVH block, CVIM, and CHIMB, respectively. $F^{L,R}_{out}, F^{L,R}_{i}$ denote the output of the $N$-th CVH Block and $i$-th CVHBlock, respectively. 

Finally, we upsample the output features to the HR size using the pixel-shuffle operation. Additionally, we incorporate a global residual path to leverage the input stereo image information to further improve the super-resolution performance. It can be expressed as:
\begin{equation}
    \begin{aligned}
        I^L_{SR} = H_{up}(F^{L}_{out})+H_{up}(I^L_{LR})=H^L_{CVHSSR}(I^L_{LR}), \\
        I^R_{SR} = H_{up}(F^{R}_{out})+H_{up}(I^R_{LR})=H^R_{CVHSSR}(I^R_{LR}),
    \end{aligned}
\end{equation}
where $H_{up}$ denotes the upsampling operation. $H_{CVHSSR}$ denotes the proposed CVHSSR network. $I^L_{SR}, I^R_{SR}$ denote the final restored left-view and right-view images, respectively.

\begin{figure}[t]
 \centering
 \includegraphics[width=\columnwidth]{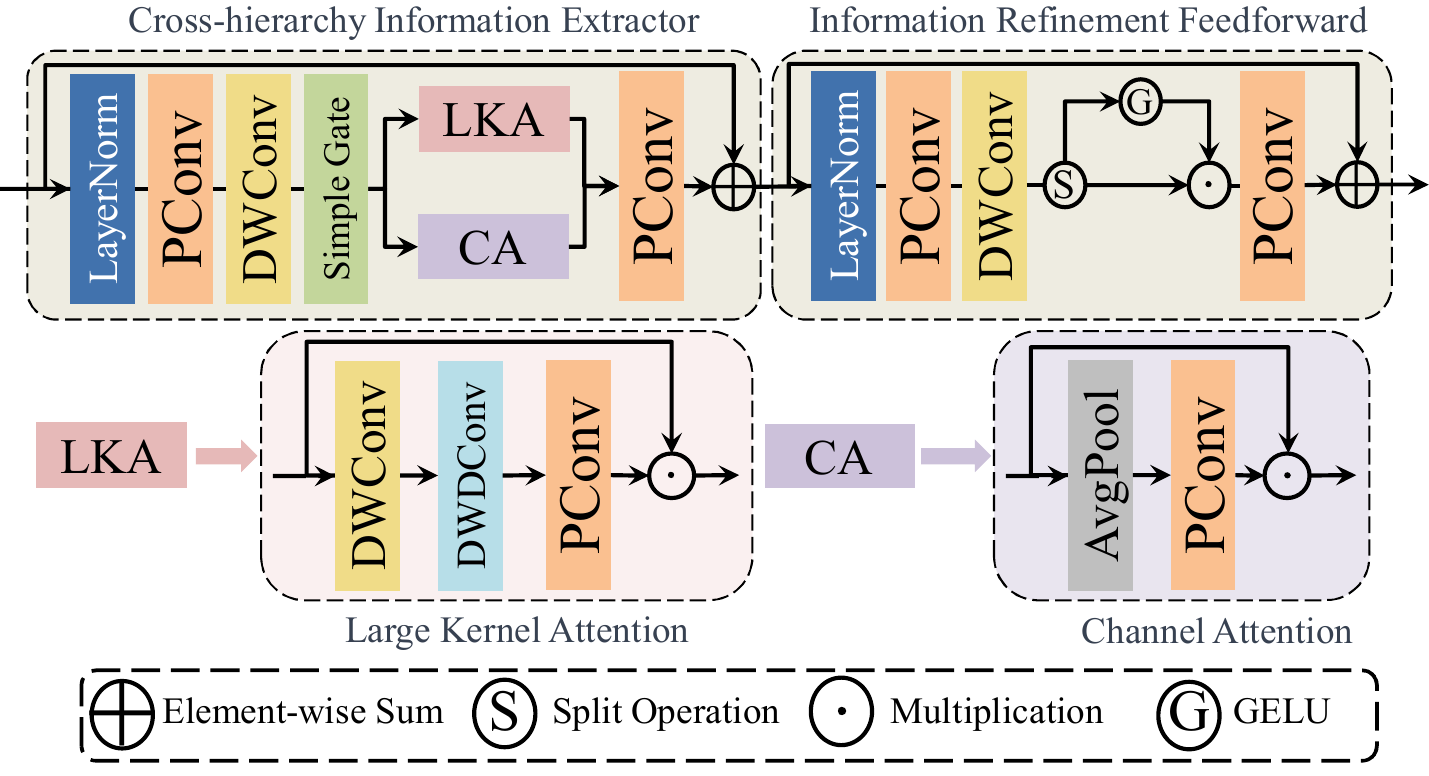}
 \caption{The architecture of our proposed cross-hierarchy information mining block (CHIMB). PConv, DWConv and DWDConv in the figure represent point-wise convolution, depth-wise convolution, and depth-wise dilation convolution, respectively.}\label{fig: CHIMB}
 \label{CHIMB}
 \end{figure}

\subsection{Cross-Hierarchy Information Mining Block}
\label{3.2}

Many existing methods for stereo image SR focus mainly on modeling cross-view information and do not adequately exploit the hierarchy information from the intra-view image. This leads to difficulties in recovering clear texture details. To address this issue, we have proposed the CHIMB as depicted in Figure \ref{CHIMB}, which is capable of effectively extracting different hierarchy information features from images. 

The CHIMB consists of two parts: (1) The cross-hierarchy information extractor (CHIE); (2) The information refinement feedforward network (IRFFN). The CHIMB incorporates both channel attention and large kernel convolution attention to capture both global and local similarity relationships. The channel attention calculates global statistics of the feature map to enhance the focus on important features. Meanwhile, the large kernel convolution attention utilizes larger kernel convolution to capture long-range dependencies in the intra-view image, thereby enhancing the attention to local information. These two attention mechanisms in combination enable the CHIE to effectively model the hierarchy information contained in the input images and accurately recover texture details. 

Given an input tensor $F_{in}\in \mathcal{R}^{H\times W\times C}$, the CHIE is formulated is:
\begin{equation}
    F_{\text{CHIE}} = W_p^0 (\mathcal{H}(\delta_{SG}(W^1_{d3} W^1_p(\text{LN}(F_{in}))))) + F_{in},
\end{equation}
where $W_p^{(\cdot)}$ is the $1\times 1$ point-wise convolution and $W_{d3}^{(\cdot)}$ is the $3\times 3$ depth-wise convolution. $F_{\text{CHIE}}$ denotes the output feature of CHIE. The $\text{LN}(\cdot)$ denotes layer normalization. We use the notation $\delta_{SG}(\cdot)$ and $\mathcal{H}(\cdot)$ to represent the SimpleGate function and the hybrid attention operation, respectively. Specifically, the SimpleGate first split the input into two features $\textbf{X}_1, \textbf{X}_2 \in \mathcal{R}^{H\times W\times C/2}$ along channel dimension. Then, it computes the output with the linear gate as $\delta_{SG}(\textbf{X}) = \textbf{X}_1\odot \textbf{X}_2$, where $\odot$ denotes element-wise multiplication. The hybrid attention $\mathcal{H}(\cdot)$ consists of two components: channel attention and large kernel convolution attention. It can be described as follows:
\begin{equation}
    \begin{aligned}
        \mathcal{H}(\textbf{X}) &= \text{LKA}(\textbf{X}) + \text{CA}(\textbf{X}),\\
        \text{CA}(\textbf{X}) &= \textbf{X} \odot (W_p H_{Avg}(\textbf{X})), \\ 
        \text{LKA}(\textbf{X}) &= \textbf{X} \odot (W_p W_{dd7} W_{d5}(\textbf{X})),
    \end{aligned}
\end{equation}
where $\text{LKA}(\cdot)$, $\text{CA}(\cdot)$ and $H_{Avg}$ denote the large kernel convolution attention, the channel attention, and the average pooling operation, respectively. $\odot$ denotes element-wise multiplication. The $W_{d5}$ and $W_{dd7}$ represent the $5\times 5$ depth-wise convolution and the $7\times 7$ depth-wise dilation convolution. 

The IRFFN in our pipeline utilizes a non-linear gate mechanism to control the flow of information, allowing each channel to focus on fine details complementary to the other levels. The IRFFN process is formulated as:
\begin{equation}
    F_{out} = W^3_p (\delta_{NG}(W^2_{d3} W^2_p(\text{LN}(F_{\text{CVIE}})))) + F_{\text{CVIE}}
\end{equation}
where $\delta_{NG}(\cdot)$ is the function of non-linear gate mechanism. Similar to SimpleGate, the non-linear gate mechanism divides the input along the channel dimension into two features $\textbf{X}_1, \textbf{X}_2 \in \mathcal{R}^{H\times W\times C/2}$. The output is then calculated by non-linear gating as $\delta_{NG}(\textbf{X}) = GELU(\textbf{X}_1)\odot \textbf{X}_2$, where $GELU(\cdot)$ denotes the activation function. The $F_{out}$ denote the output of CHIMB.
\begin{figure}[t]
 \centering
 \includegraphics[width=\columnwidth]{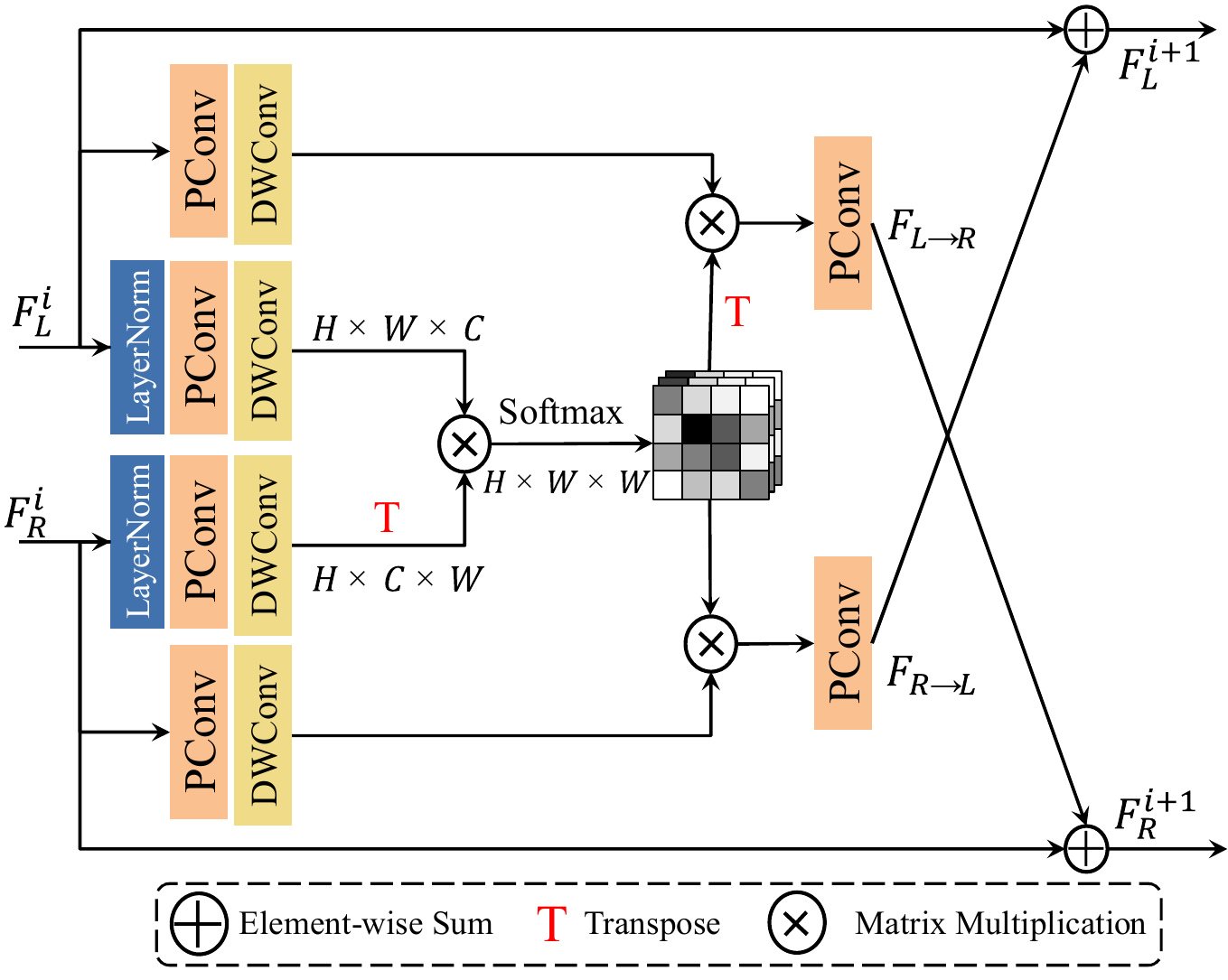}
 \caption{The architecture of our proposed cross-view interaction modules (CVIM). PConv and DWConv in the figure represent point-wise convolution, and depth-wise convolution, respectively.}\label{fig: CVIM}
 \end{figure}
 
\subsection{Cross-View Interaction Module}
\label{3.3}

The details of the proposed CVIM as shown in Figure \ref{fig: CVIM}. This method employs Scaled DotProduct Attention \cite{AiAyn}, which involves calculating the dot product between the query and keys, followed by the application of a softmax function to generate weights assigned to the corresponding values:
\begin{equation}
    Attention(\textbf{Q},\textbf{K},\textbf{V}) = \text{Softmax}(\textbf{Q}\textbf{K}^T / {\sqrt{C}})\textbf{V},
\end{equation}
where $\textbf{Q} \in \mathcal{R}^{H\times W\times C}$ is query matrix projected by source intra-view feature (e.g. left-view), and $\textbf{K},\textbf{V}\in \mathcal{R}^{H\times W\times C}$ are key, value matrices projected by target intra-view feature (e.g. right-view). Here, $H, W, C$ represents the height, width, and number of channels of the feature map. 

The CVIM adopts a novel cross-view attention mechanism that integrates information from both left and right view images to generate cross-view attention maps. This approach allows for the exploitation of the distinctive information present in each view, leading to more effective feature fusion and better restoration results. In detail, given the input stereo intra-view features $F_L^i, F_R^i \in \mathcal{R}^{H\times W\times C}$, we can get the cross-view fusion features $F_{L\rightarrow R}$ as follows:
\begin{align}
    \textbf{Q}_L &= W_d^{Q_L}W_p^{Q_L}(LN(F_L^i)),\\
    \textbf{K}_R &= W_d^{K_R}W_p^{K_R}(LN(F_R^i)),\\
     \textbf{V}_R &= W_d^{V_R}W_p^{V_R}F_R^i, \\
     F_{L\rightarrow R} &= W_p^R Attention_{L\rightarrow R}(\textbf{Q}_L,\textbf{K}_R,\textbf{V}_R),
\end{align}
where $W_p^{(\cdot)}$ is the $1\times 1$ point-wise convolution. $W_d^{(\cdot)}$ is the $3\times 3$ depth-wise convolution. It offers feature refinement from both channel and spatial perspectives. 

With an analogous argument, we can obtain the cross-view fusion features $F_{R\rightarrow L}$ as follows:
\begin{align}
    \textbf{Q}_R &= W_d^{Q_R}W_p^{Q_R}(LN(F_R^i)),\\
    \textbf{K}_L &= W_d^{K_L}W_p^{K_L}(LN(F_L^i)),\\
     \textbf{V}_L &= W_d^{V_L}W_p^{V_L}F_L^i, \\
     F_{R\rightarrow L} &= W_p^L Attention_{R\rightarrow L}(\textbf{Q}_R,\textbf{K}_L,\textbf{V}_L),
\end{align}
Finally, the interacted cross-view information $F_{L\rightarrow R}$, $F_{R\rightarrow L}$ and intra-view information $F_L^i$, $F_R^i$ are fused by element-wise addition:
\begin{align}
    F_L^{i+1} = \gamma_L F_{L\rightarrow R} + F_L^i, \\
    F_R^{i+1} = \gamma_R F_{R\rightarrow L} + F_R^i, 
\end{align}
where $\gamma_L$ and $\gamma_R$ are trainable channel-wise scales and initialized with zeros for stabilizing training.

\begin{table*}
\caption{Quantitative results achieved by different methods on the KITTI2012, KITTI2015, Middlebury, and Flickr1024 datasets. Params represents the number of parameters of the networks. Here, PSNR/SSIM values achieved on both the left images (i.e., \textit{Left}) and a pair of stereo images (i.e., $(Left + Right)/2$) are reported. The best and second best results are \textcolor{red}{red} and \textcolor{blue}{blue}.}
\label{tab:my-table}
\centering
\resizebox{16cm}{!}{%
\begin{tabular}{llc|ccc|cccc}
\bottomrule[1.2pt]
\multirow{2}{*}{Method} & \multicolumn{1}{c}{\multirow{2}{*}{Scale}} & \multirow{2}{*}{Params} & \multicolumn{3}{c|}{$Left$}                  & \multicolumn{4}{c}{$(Left+Right)/2$}                        \\ \cline{4-10} 
                        & \multicolumn{1}{c}{}                       &                         & KITTI2012    & KITTI2015    & Middlebury   & KITTI2012    & KITTI2015    & Middlebury   & Flickr1024   \\ \hline
VDSR \cite{VDSR}        & $\times 2$                                 & 0.66M                   & 30.17/0.9062 & 28.99/0.9038 & 32.66/0.9101 & 30.30/0.9089 & 29.78/0.9150 & 32.77/0.9102 & 25.60/0.8534 \\
EDSR \cite{EDSR}        & $\times 2$                                 & 38.6M                   & 30.83/0.9199 & 29.94/0.9231 & 34.84/0.9489 & 30.96/0.9228 & 30.73/0.9335 & 34.95/0.9492 & 28.66/0.9087 \\
RDN  \cite{RDN}         & $\times 2$                                 & 22.0M                   & 30.81/0.9197 & 29.91/0.9224 & 34.85/0.9488 & 30.94/0.9227 & 30.70/0.9330 & 34.94/0.9491 & 28.64/0.9084 \\
RCAN \cite{RCAN}        & $\times 2$                                 & 15.3M                   & 30.88/0.9202 & 29.97/0.9231 & 34.80/0.9482 & 31.02/0.9232 & 30.77/0.9336 & 34.90/0.9486 & 28.63/0.9082 \\
StereoSR \cite{Jeon}& $\times 2$                                 & 1.08M                   & 29.42/0.9040 & 28.53/0.9038 & 33.15/0.9343 & 29.51/0.9073 & 29.33/0.9168 & 33.23/0.9348 & 25.96/0.8599 \\
PASSRnet \cite{PASSRNet}& $\times 2$                                 & 1.37M                   & 30.68/0.9159 & 29.81/0.9191 & 34.13/0.9421 & 30.81/0.9190 & 30.60/0.9300 & 34.23/0.9422 & 28.38/0.9038 \\
IMSSRnet \cite{IMSSRNet}& $\times 2$                                 & 6.84M                   & 30.90/-      & 29.97/-      & 34.66/-      & 30.92/-      & 30.66/-      & 34.67/-      & -/-          \\
iPASSR  \cite{iPASSR}   & $\times 2$                                 & 1.37M                   & 30.97/0.9210 & 30.01/0.9234 & 34.41/0.9454 & 31.11/0.9240 & 30.81/0.9340 & 34.51/0.9454 & 28.60/0.9097 \\
SSRDE-FNet \cite{SSRDE} & $\times 2$                                 & 2.10M                   & 31.08/0.9224 & 30.10/0.9245 & 35.02/0.9508 & 31.23/0.9254 & 30.90/0.9352 & 35.09/0.9511 & 28.85/0.9132 \\
PFT-SSR \cite{PFT}      & $\times 2$                                 & -                       & 31.15/0.9166 & 30.16/0.9187 & 35.08/0.9516 & 31.29/0.9195 & 30.96/0.9306 & 35.21/0.9520 & 29.05/0.9049 \\
SwinFIR-T \cite{swinfir}& $\times 2$                                 & 0.89M                   & 31.09/0.9226 & 30.17/0.9258 & 35.00/0.9491 & 31.22/0.9254 & 30.96/0.9359 & 35.11/0.9497 & 29.03/0.9134 \\
NAFSSR-T \cite{nafssr} & $\times 2$                                  & 0.45M                   & 31.12/0.9224 & 30.19/0.9253 & 34.93/0.9495 & 31.26/0.9254 & 30.99/0.9355 & 35.01/0.9495 & 28.94/0.9128 \\
NAFSSR-S \cite{nafssr} & $\times 2$                                  & 1.54M                   & 31.23/0.9236 & 30.28/0.9266 & 35.23/0.9515 & 31.38/0.9266 & 31.08/0.9367 & 35.30/0.9514 & 29.19/0.9160 \\
NAFSSR-B \cite{nafssr} & $\times 2$                                  & 6.77M                   & \textcolor{blue}{31.40}/\textcolor{blue}{0.9254} & \textcolor{blue}{30.42}/\textcolor{blue}{0.9282} & \textcolor{blue}{35.62}/\textcolor{blue}{0.9545} & \textcolor{blue}{31.55}/\textcolor{blue}{0.9283} & \textcolor{blue}{31.22}/\textcolor{blue}{0.9380} & \textcolor{blue}{35.68}/\textcolor{blue}{0.9544} & \textcolor{blue}{29.54}/\textcolor{blue}{0.9204} \\ \hline
CVHSSR-T (Ours)        & $\times 2$                                  & 0.66M                   & 31.31/0.9250 & 30.33/0.9277 &         35.41/0.9533 & 31.46/0.9280 & 31.13/0.9377 &35.47/0.9532 & {29.26/0.9180}    \\
CVHSSR-S (Ours)        & $\times 2$                                 &2.22M                    & \textcolor{red}{31.42}/\textcolor{red}{0.9262} & \textcolor{red}{30.42}/\textcolor{red}{0.9287} &         \textcolor{red}{35.73}/\textcolor{red}{0.9551} & \textcolor{red}{31.57}/\textcolor{red}{0.9291} & \textcolor{red}{31.22}/\textcolor{red}{0.9385} &\textcolor{red}{35.78}/\textcolor{red}{0.9550} & \textcolor{red}{29.56}/\textcolor{red}{0.9216}  \\ \hline \hline
VDSR \cite{VDSR}       & $\times 4$                                 & 0.66M                   & 25.54/0.7662 & 24.68/0.7456 & 27.60/0.7933 & 25.60/0.7722 & 25.32/0.7703 & 27.69/0.7941 & 22.46/0.6718 \\
EDSR \cite{EDSR}       & $\times 4$                                 & 38.9M                   & 26.26/0.7954 & 25.38/0.7811 & 29.15/0.8383 & 26.35/0.8015 & 26.04/0.8039 & 29.23/0.8397 & 23.46/0.7285 \\
RDN \cite{RDN}         & $\times 4$                                 & 22.0M                   & 26.23/0.7952 & 25.37/0.7813 & 29.15/0.8387 & 26.32/0.8014 & 26.04/0.8043 & 29.27/0.8404 & 23.47/0.7295 \\
RCAN \cite{RCAN}       & $\times 4$                                 & 15.4M                   & 26.36/0.7968 & 25.53/0.7836 & 29.20/0.8381 & 26.44/0.8029 & 26.22/0.8068 & 29.30/0.8397 & 23.48/0.7286 \\
StereoSR \cite{Jeon}   & $\times 4$                                 & 1.42M                   & 24.49/0.7502 & 23.67/0.7273 & 27.70/0.8036 & 24.53/0.7555 & 24.21/0.7511 & 27.64/0.8022 & 21.70/0.6460 \\
PASSRnet \cite{PASSRNet}& $\times 4$                                 & 1.42M                   & 26.26/0.7919 & 25.41/0.7772 & 28.61/0.8232 & 26.34/0.7981 & 26.08/0.8002 & 28.72/0.8236 & 23.31/0.7195 \\
SRRes+SAM \cite{Ressam} & $\times 4$                                 & 1.73M                   & 26.35/0.7957 & 25.55/0.7825 & 28.76/0.8287 & 26.44/0.8018 & 26.22/0.8054 & 28.83/0.8290 & 23.27/0.7233 \\
IMSSRnet \cite{IMSSRNet}& $\times 4$                                 & 6.89M                   & 26.44/-      & 25.59/-      & 29.02/-      & 26.43/-      & 26.20/-      & 29.02/-      & -/-          \\
iPASSR \cite{iPASSR}    & $\times 4$                                 & 1.42M                   & 26.47/0.7993 & 25.61/0.7850 & 29.07/0.8363 & 26.56/0.8053 & 26.32/0.8084 & 29.16/0.8367 & 23.44/0.7287 \\
SSRDE-FNet \cite{SSRDE} & $\times 4$                                 & 2.24M                   & 26.61/0.8028 & 25.74/0.7884 & 29.29/0.8407 & 26.70/0.8082 & 26.43/0.8118 & 29.38/0.8411 & 23.59/0.7352 \\
PFT-SSR \cite{PFT}      & $\times 4$                                 & -                       & 26.64/0.7913 & 25.76/0.7775 & 29.58/0.8418 & 26.77/0.7998 & 26.54/0.8083 & 29.74/0.8426 & 23.89/0.7277 \\
SwinFIR-T \cite{swinfir}& $\times 4$                                 & 0.89M                   & 26.59/0.8017 & 25.78/0.7904 & 29.36/0.8409 & 26.68/0.8081 & 26.51/0.8135 & 29.48/0.8426 & 23.73/0.7400 \\
NAFSSR-T \cite{nafssr}  & $\times 4$                                 & 0.46M                   & 26.69/0.8045 & 25.90/0.7930 & 29.22/0.8403 & 26.79/0.8105 & 26.62/0.8159 & 29.32/0.8409 & 23.69/0.7384 \\
NAFSSR-S \cite{nafssr}  & $\times 4$                                 & 1.56M                   & 26.84/0.8086 & 26.03/0.7978 & 29.62/0.8482 & 26.93/0.8145 & 26.76/0.8203 & 29.72/0.8490 & 23.88/0.7468 \\ 
NAFSSR-B \cite{nafssr}  & $\times 4$                                 & 6.80M                   & \textcolor{blue}{26.99}/\textcolor{blue}{0.8121} & \textcolor{red}{26.17}/\textcolor{blue}{0.8020} & \textcolor{blue}{29.94}/\textcolor{blue}{0.8561} & \textcolor{blue}{27.08}/\textcolor{blue}{0.8181} & \textcolor{red}{26.91}/\textcolor{blue}{0.8245} & \textcolor{blue}{30.04}/\textcolor{blue}{0.8568} & \textcolor{blue}{24.07}/\textcolor{blue}{0.7551} \\ \hline
CVHSSR-T (Ours)         & $\times 4$                                 & 0.68M                   & 26.88/0.8105 & 26.03/0.7991 &         29.62/0.8496 & 26.98/0.8165 & 26.78/0.8218 &29.74/0.8505 & {23.89/0.7484} \\
CVHSSR-S (Ours)         & $\times 4$                                 & 2.24M                   & \textcolor{red}{27.00}/\textcolor{red}{0.8139} & \textcolor{blue}{26.15}/\textcolor{red}{0.8033} &        \textcolor{red}{29.94}/\textcolor{red}{0.8577} & \textcolor{red}{27.10}/\textcolor{red}{0.8199} & \textcolor{blue}{26.90}/\textcolor{red}{0.8258} &\textcolor{red}{30.05}/\textcolor{red}{0.8584} & \textcolor{red}{24.08}/\textcolor{red}{0.7570} \\ \toprule[1.2pt]
\end{tabular}%
}
\label{Quantitative Compare}
\end{table*}

\begin{figure*}
	\centering %15cm
	\resizebox{13.cm}{!}{\includegraphics[]{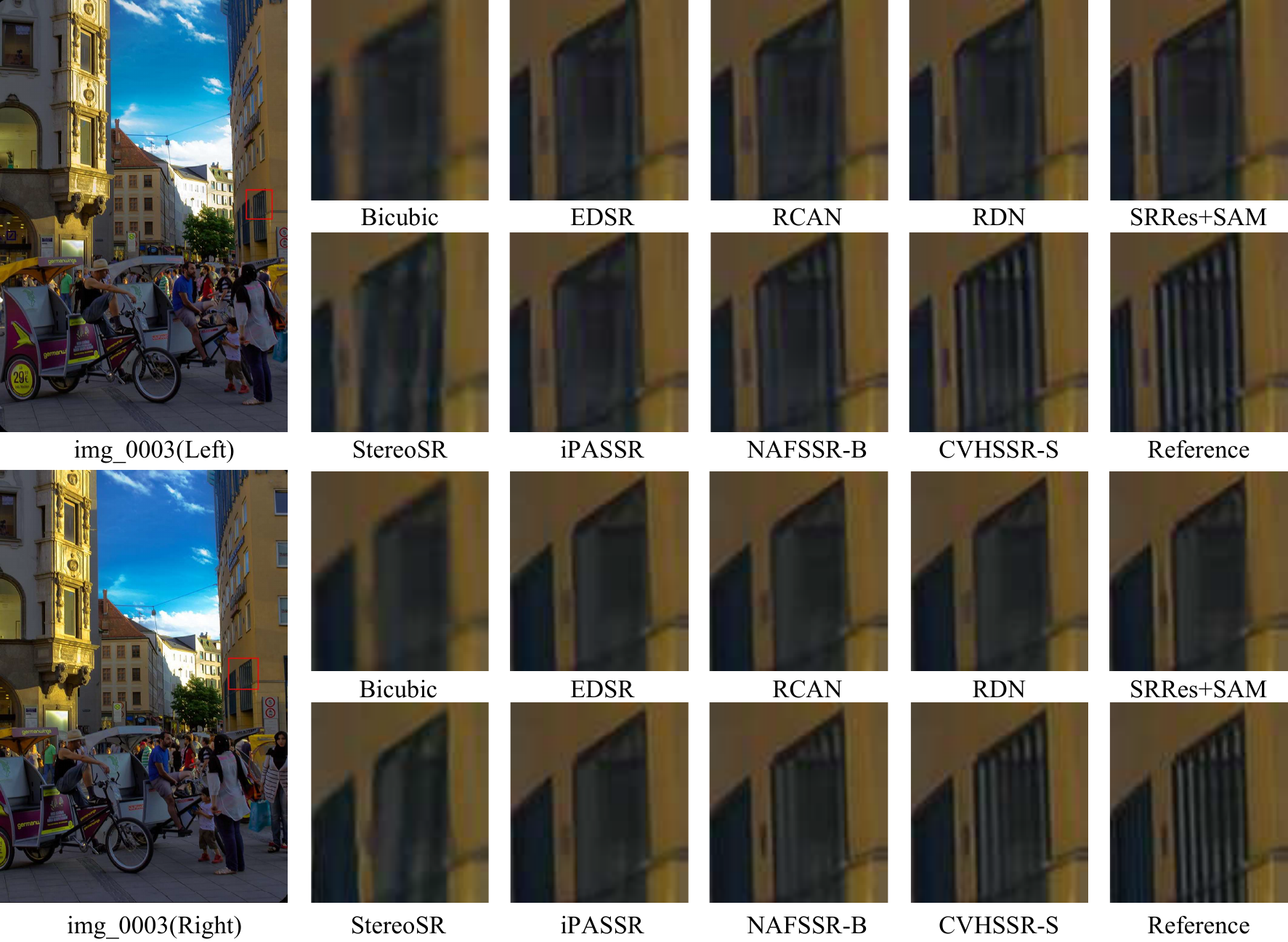}}\\
	\caption{Visual results ($\times 4$) achieved by different methods on the Flickr1024 \cite{Flickr1024} dataset.}
	\label{cpimageflickr}
\end{figure*}

\begin{figure*}
	\centering %16cm
	\resizebox{13.5cm}{!}{\includegraphics[]{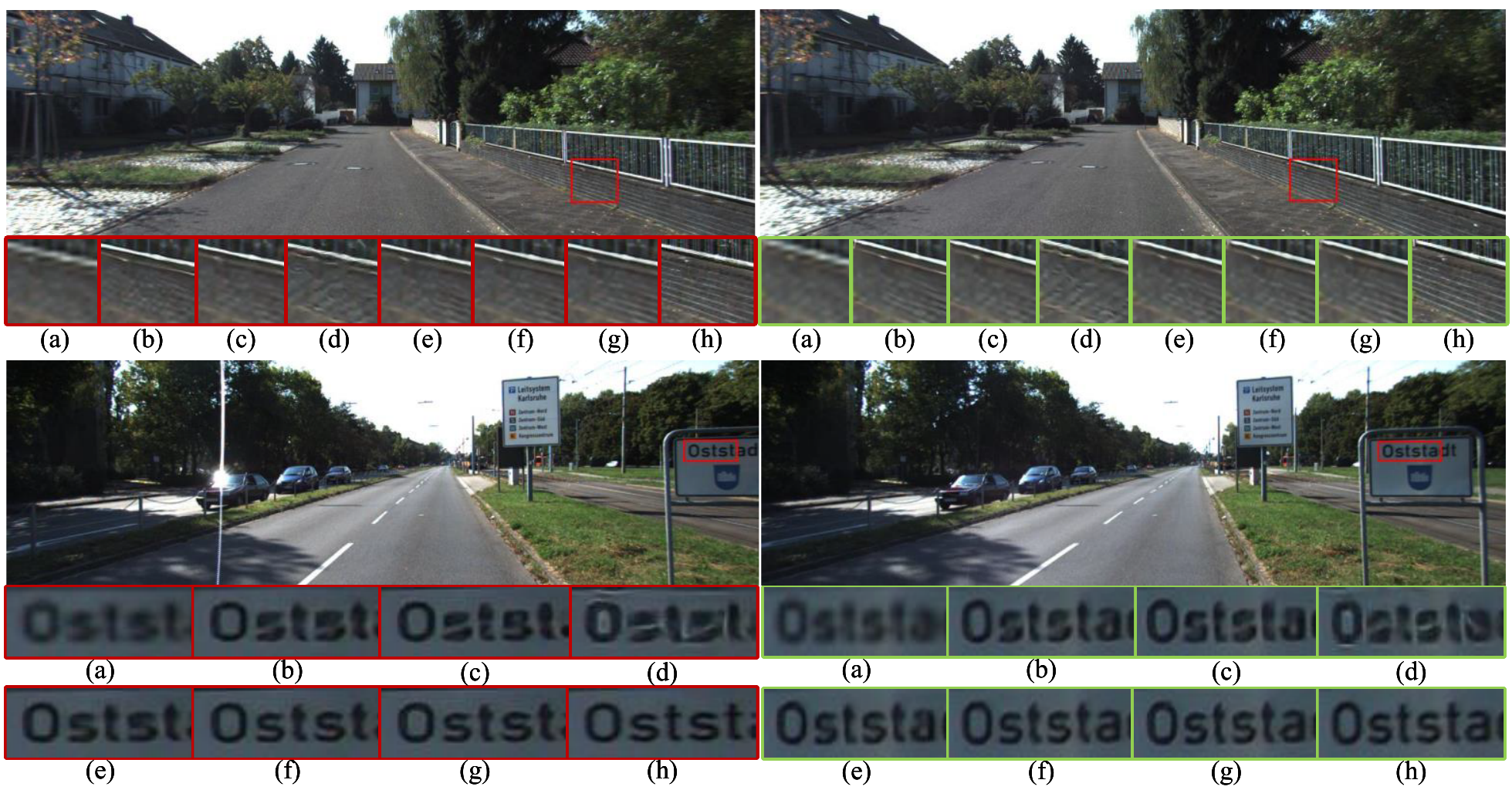}}\\
	\caption{Visual results ($\times 4$) achieved by different methods on the KITTI 2012 \cite{KITTI2012} (top) and KITTI 2015 \cite{KITTI2015} (bottom) dataset. The images with red and green borders represent the left and right views respectively. (a) Bicubic. (b) SRRes+SAM \cite{Ressam}. (c) RDN \cite{RDN}. (d) StereoSR \cite{Jeon}. (e) iPASSR \cite{iPASSR}. (f) NAFSSR-B \cite{nafssr}. (g) CVHSSR (Ours). (h) Reference.}
	\label{cpimagekitti}
\end{figure*}

\subsection{Loss Function}
\label{3.4}
As image super-resolution tasks are primarily focused on restoring high-frequency details, we leverage both spatial and frequency domain losses to jointly guide our network in effectively recovering clear and sharp high-frequency textures. Specifically, given an input two view LR image $I_{LR}^{L,R}$, the proposed model predicts HR stereo image denote $I_{SR}^{L,R}$. We optimize our CVHSSR with the following loss function:
\begin{equation}
        \mathcal{L}_{total} = \mathcal{L}_{MSE}(I_{SR}^{L,R}, I_{HR}^{L,R}) + \lambda *\mathcal{L}_{FC}(I_{SR}^{L,R}, I_{HR}^{L,R}),\label{con:loss}
\end{equation}
where $I_{HR}^{L,R}$ represents the left-view and right-view HR image, and $\mathcal{L}_{MSE}$ is the MSE loss:
\begin{equation}
    \mathcal{L}_{MSE}=\dfrac{1}{N}\sum_{i=1}^{N}||I_{HR}^{L,R}-I_{SR}^{L,R}||^2,
\end{equation}
In addition, $\mathcal{L}_{FC}$ is the frequency Charbonnier loss, defined as:
\begin{equation}
    \mathcal{L}_{FC}=\dfrac{1}{N}\sum_{i=1}^{N}\sqrt{||FFT(I_{HR}^{L,R})-FFT(I_{SR}^{L,R})||^2+\epsilon^2},
\end{equation}
with constant $\epsilon$ emiprically set to $10^{-3}$ for all the experiments. $FFT(\cdot)$ denotes a fast Fourier transform. The parameter $\lambda$ in Eq. (\ref{con:loss}) is a hyper-parameter used to control the composition of the frequency Charbonnier loss function. The parameter $\lambda$ is set to $0.01$ for all the experiments. More training details of our method are presented in Section \ref{sec:experiments}.

%------------------------------------------------------------------------
\section{Experiments}
\label{sec:experiments}

% \begin{figure*}
% \vspace{-0.6cm}
% \setlength{\abovecaptionskip}{-0.35cm} 
% \setlength{\belowcaptionskip}{-0.8cm}%调整caption与下文的距离
% 	\centering
% 	\includegraphics[scale=0.7,trim={1cm 5cm 0.5cm 0.5cm},clip]{network_1.pdf}\\
% 	\caption{Qualitative comparison with the leading algorithms: SRCNN \cite{SRCNN}, VDSR \cite{VDSR}, CARN \cite{CARN}, IDN \cite{IDN}, IMDN \cite{IMDN}, PAN \cite{PAN}, and RFDN \cite{RFDN} on $\times$4 task. From the figure, we can see that our method can generate finer details of the image and achieve outstanding performance.}
% 	\label{imgcp}
% \end{figure*}

\subsection{Implementation Details}
In this section, we provide a detailed description of the experimental setting, including the datasets, the evaluation metrics, and the training configurations.

\textbf{Dataset.} Following the previous methods \cite{iPASSR, nafssr, swinfir}, we employ the training and validation datasets provided by the Flickr1024 \cite{Flickr1024}. To be specific, we employ 800 stereo images as the training data, and 112 stereo images as the validation data. We augment the training data with random horizontal, flips, rotations, and RGB channel shuffle. For testing, we use four benchmark datasets: KITTI 2012 \cite{KITTI2012}, KITTI 2015 \cite{KITTI2015}, Middlebury \cite{Middlebury}, and Flickr1024 \cite{Flickr1024}.

\textbf{Evaluation metrics.} We adopted peak signal-to-noise ratio (PSNR) and structural similarity (SSIM) as quantitative metrics for evaluation, which are calculated in the RGB color space between a pair of stereo images (i.e.,$(Left+Right)/2$).

\textbf{Model Setting.}
The numbers of the CHIMB blocks and feature channels are flexible and configurable. We construct two CVHSSR networks of varying sizes, which we named CVHSSR-T (Tiny) and CVHSSR-S (Small) by adjusting the number of channels and blocks. Specifically, the number of channels and blocks for CVHSSR-T are set to 48 and 16 respectively. The number of channels and blocks for CVHSSR-S are set to 64 and 32 respectively. 

\textbf{Training Settings.} Our network was optimized using the Lion method \cite{chen2023symbolic} with $\beta_1$=0.9, $\beta_2$=0.999, and a batch size of 8. Our CVHSSR was implemented in PyTorch on a PC with four Nvidia RTX 3090 GPUs. The learning rate was initially set to $5\times10^{-4}$ and decayed the learning rate with the cosine strategy. We trained this model for 200,000 iterations. To alleviate the overfitting issue, we use stochastic depth \cite{huang2016deep} with 0.1 and 0.2 probability for CVHSSR-S and CVHSSR-B, respectively. Moreover, we also use Test-time Local Converter (TLC) \cite{TLC} to further improve the model performance. The TLC method mainly aims to reduce the discrepancy between the distribution of global information during training and inference by converting global operations into local operations at inference.

\subsection{Comparisons with State-of-the-art Methods}

In this section, we conduct a comparative analysis between our CVHSSR (with 2 different variations) and existing super-resolution (SR) methods. The comparison involved SISR methods such as VDSR \cite{VDSR}, EDSR \cite{EDSR}, RDN \cite{RDN}, and RCAN \cite{RCAN}, as well as stereo image SR methods like StereoSR \cite{Jeon}, PASSRnet \cite{PASSRNet}, SRRes+SAM \cite{Ressam}, IMSSRnet \cite{IMSSRNet}, iPASSR \cite{iPASSR}, SSRDE-FNet \cite{SSRDE}, NAFSSR \cite{nafssr}, SwinFIR \cite{swinfir}, and PFT-SSR \cite{PFT}. All these methods were trained on the same datasets as ours, and their PSNR and SSIM scores were evaluated and reported by \cite{nafssr}.

\textbf{Quantitative Evaluations.} 
We present a comparative evaluation of our proposed CVHSSR against existing stereo SR methods at $\times 2$ and $\times 4$ upscaling factors, as summarized in Table \ref{Quantitative Compare}. Notably, even our smallest CVHSSR-T model outperforms the NAFSSR-S method on all datasets while utilizing 60\% fewer parameters. Moreover, our CVHSSR-S model achieves better results than the NAFSSR-B method while requiring 70\% fewer parameters. Specifically, our CVHSSR-S method is 0.19 dB and 0.48 dB higher than NAFSSR-S and SRRDE-FNet, respectively, on the $\times 4$ Flickr1024 dataset for the same amount of parameters. These results demonstrate the effectiveness of our proposed method and its superiority over existing methods in stereo image SR tasks.

\textbf{Visual Comparison.} 
In Figures \ref{cpimageflickr} and \ref{cpimagekitti}, we present comparative results of $\times 4$ stereo SR obtained by different stereo SR methods on the Flickr1024 \cite{Flickr1024}, KITTI2012 \cite{KITTI2012}, and KITTI2015 \cite{KITTI2015} datasets. The visual comparison of reconstructed images demonstrates that our proposed CVHSSR-S method achieves sharper and more accurate texture details compared to the NAFSSR-B method which still suffers from over-smoothing fine textures. This validates the superiority and effectiveness of our proposed CVHSSR method.

\subsection{Ablation Study}
In this section, we conduct a set of ablation experiments to evaluate the performance of each proposed module. The evaluation is performed on the Flickr1024 \cite{Flickr1024} validation dataset.

\textbf{Effectiveness of Each Operation.} 
To further substantiate the effectiveness of our proposed module, a series of ablation experiments were conducted and the results are presented in Table \ref{tab:components}. Initially, the NAFSSR was used as the baseline, and subsequently, the corresponding module was modified continuously to verify the efficacy of the proposed module. As depicted in the table, LKA provided a performance improvement of 0.08 dB to the baseline due to its larger receptive field. However, merely enlarging the receptive field of the network does not fully exploit the hierarchical relationship of the intra-view. Hence, our proposed CHIE provided a superior performance improvement of 0.1 dB to the network. Compared to the simple FFN in NAFSSR \cite{nafssr}, our proposed IRFFN more effectively regulated the information flow. Additionally, our proposed CVIM demonstrated a superior ability to fuse similar information from various perspectives and enhance network performance, compared to the traditional PAM \cite{iPASSR}. These comparisons unequivocally underscore the effectiveness of our proposed methods.

\begin{table}
\caption{Ablation studies of different components. We report the PSNR (dB) values on Flickr 1024 validation datasets ($\times 4$).}
\label{tab:components}
\centering
\resizebox{7cm}{!}{%
\begin{tabular}{c|ccccc}
\bottomrule[1.2pt]
         & 1      & 2     & 3      & 4      & 5      \\ \hline
Baseline & \ding{52}    & \ding{52}   & \ding{52}    & \ding{52}    & \ding{52}    \\
LKA      & \ding{56}    & \ding{52}   & \ding{56}    & \ding{56}    & \ding{56}    \\
CHIE     & \ding{56}    & \ding{56}   & \ding{52}    & \ding{52}    & \ding{52}    \\
IRFFN    & \ding{56}    & \ding{56}   & \ding{56}    & \ding{52}    & \ding{52}    \\
CVIM     & \ding{56}    & \ding{56}   & \ding{56}    & \ding{56}    & \ding{52}    \\ \hline
PSNR     & 23.59  & 23.67 & 23.69  & 23.70   & \textbf{23.72}  \\
SSIM     & 0.7345 & 0.739 & 0.7399 & 0.7402 & \textbf{0.7413} \\
$\Delta$ PSNR    & 0      & 0.08  & 0.10    & 0.11   & \textbf{0.13}   \\ \toprule[1.2pt]
\end{tabular}%
}
\end{table}

\begin{figure}
	\centering
    \resizebox{\columnwidth}{!}{\includegraphics[]{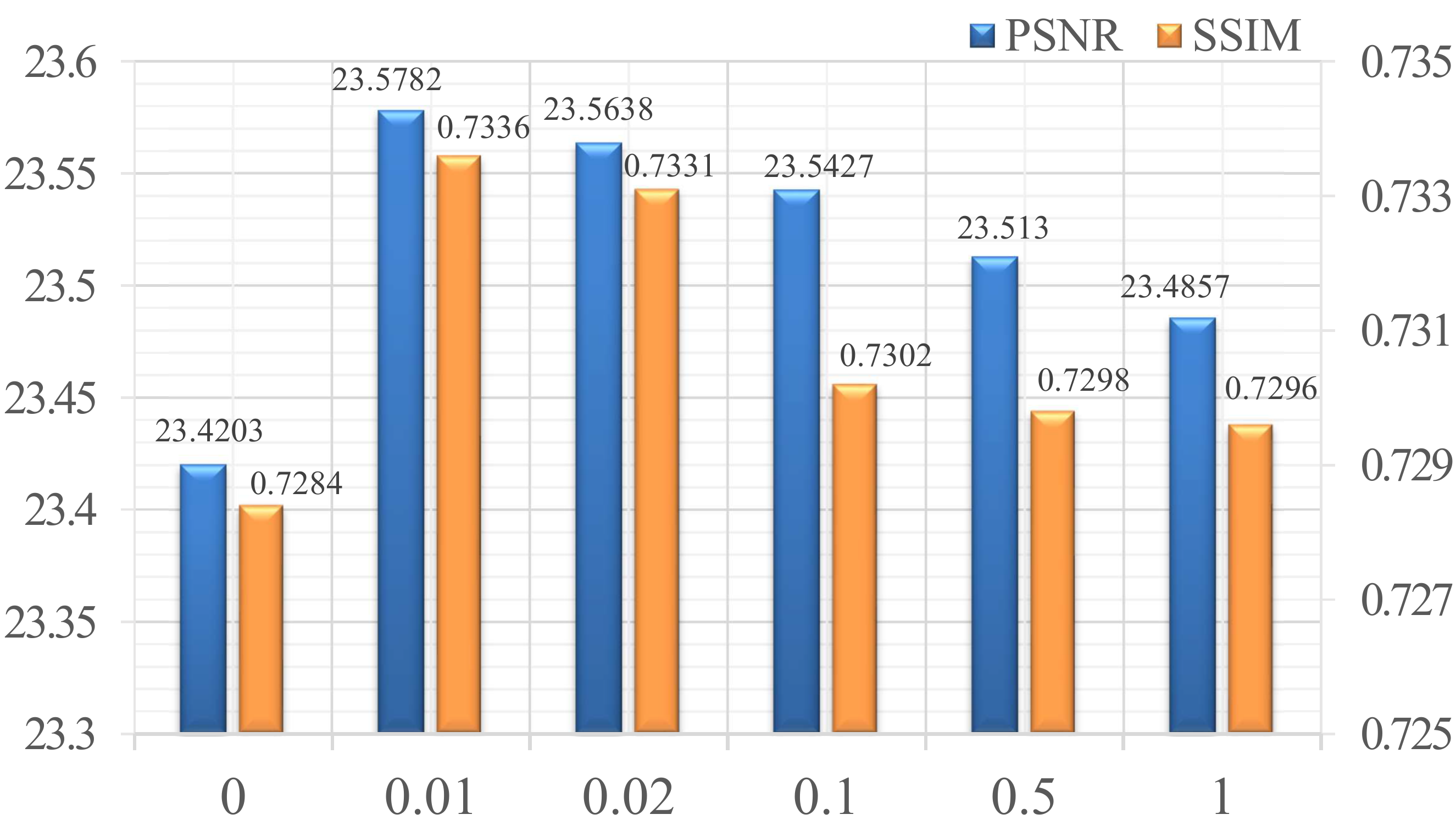}}
	\caption{Study on the influence of $\lambda$ in loss function. The PSNR performance on the Flickr 1024 dataset ($\times 4$). When $\lambda=0.01$, the model achieves the best results.}
	\label{Loss}
\end{figure}
\textbf{Effectiveness of $\lambda$ in loss function.}
To evaluate the influence of different values of $\lambda$ in the loss function, a series of experiments are conducted in this section. The hyperparameter $\lambda$ is utilized to balance the trade-off between the MSE loss function and the frequency Charbonnier loss function. We conducted a range of empirical experiments with six different $\lambda$ values within the range of [0,1], based on previous experience. The impact of different $\lambda$ values on the model performance is illustrated in Figure \ref{Loss}. The results demonstrate that the network achieves optimal performance at $\lambda=0.01$. 

\subsection{NTIRE Stereo Image SR Challenge}
We have submitted the results obtained from our proposed approach to the NTIRE 2023 Stereo Image Super-Resolution Challenge. To enhance the potential performance of our method, we have increased the depth and width of the CVHSSR-Based model. During test-time, we employed self-ensemble and model ensemble strategies. The final submission achieves 24.114 dB PSNR on the validation set and achieves 23.742 dB PSNR on the test set. We won 8th, 5th, and 4th on track 1 Fidelity \& Bicubic, track 2 Perceptual \& Bicubic, and tack 3 Fidelity \& Realistic, respectively.
%------------------------------------------------------------------------
\section{Conclusion}
In this paper, we present an efficient stereo image SR method named Cross-View Hierarchy Network (CVHSSR). In particular, we design a cross-hierarchy information mining block that efficiently extracts similar features from intra-views by leveraging the hierarchical relationships of the features. Additionally, we introduce a cross-view interaction module to effectively convey mutual information between different views. The integration of these two modules enables our proposed network to achieve superior performance with fewer parameters. Comprehensive experimental evaluations demonstrate that our proposed CVHSSR method outperforms current state-of-the-art models in stereo image SR.

\section*{Acknowledgments} 
This work was funded by Science and Technology Project of Guangzhou (202103010003), Science and Technology Project in key areas of Foshan (2020001006285), Natural Science Foundation of Guangdong Province (2019A1515011041), Xijiang Innovation Team Project (XJCXTD3-2019-04B).

%%%%%%%%% REFERENCES
{\small
\bibliographystyle{ieee_fullname}
\bibliography{egbib}

\begin{thebibliography}{10}\itemsep=-1pt

\bibitem{nafnet}
Liangyu Chen, Xiaojie Chu, Xiangyu Zhang, and Jian Sun.
\newblock Simple baselines for image restoration.
\newblock In {\em Computer Vision--ECCV 2022: 17th European Conference, Tel
  Aviv, Israel, October 23--27, 2022, Proceedings, Part VII}, pages 17--33.
  Springer, 2022.

\bibitem{chen2023symbolic}
Xiangning Chen, Chen Liang, Da Huang, Esteban Real, Kaiyuan Wang, Yao Liu, Hieu
  Pham, Xuanyi Dong, Thang Luong, Cho-Jui Hsieh, et~al.
\newblock Symbolic discovery of optimization algorithms.
\newblock {\em arXiv preprint arXiv:2302.06675}, 2023.

\bibitem{TLC}
Xiaojie Chu, Liangyu Chen, , Chengpeng Chen, and Xin Lu.
\newblock Improving image restoration by revisiting global information
  aggregation.
\newblock {\em arXiv preprint arXiv:2112.04491}, 2021.

\bibitem{nafssr}
Xiaojie Chu, Liangyu Chen, and Wenqing Yu.
\newblock Nafssr: stereo image super-resolution using nafnet.
\newblock In {\em Proceedings of the IEEE/CVF Conference on Computer Vision and
  Pattern Recognition}, pages 1239--1248, 2022.

\bibitem{SSRDE}
Qinyan Dai, Juncheng Li, Qiaosi Yi, Faming Fang, and Guixu Zhang.
\newblock Feedback network for mutually boosted stereo image super-resolution
  and disparity estimation.
\newblock In {\em Proceedings of the 29th ACM International Conference on
  Multimedia}, pages 1985--1993, 2021.

\bibitem{SAN}
T. {Dai}, J. {Cai}, Y. {Zhang}, S. {Xia}, and L. {Zhang}.
\newblock Second-order attention network for single image super-resolution.
\newblock In {\em 2019 IEEE/CVF Conference on Computer Vision and Pattern
  Recognition (CVPR)}, pages 11057--11066, 2019.

\bibitem{DaiRBAN}
Tao Dai, Hua Zha, Yong Jiang, and Shu-Tao Xia.
\newblock Image super-resolution via residual block attention networks.
\newblock In {\em Proceedings of the IEEE/CVF International Conference on
  Computer Vision Workshops}, pages 0--0, 2019.

\bibitem{SRCNN}
C. {Dong}, C.~C. {Loy}, K. {He}, and X. {Tang}.
\newblock Image super-resolution using deep convolutional networks.
\newblock {\em IEEE Transactions on Pattern Analysis and Machine Intelligence},
  38(2):295--307, 2016.

\bibitem{KITTI2012}
Andreas Geiger, Philip Lenz, and Raquel Urtasun.
\newblock Are we ready for autonomous driving? the kitti vision benchmark
  suite.
\newblock In {\em 2012 IEEE conference on computer vision and pattern
  recognition}, pages 3354--3361. IEEE, 2012.

\bibitem{PFT}
Hansheng Guo, Juncheng Li, Guangwei Gao, Zhi Li, and Tieyong Zeng.
\newblock Pft-ssr: Parallax fusion transformer for stereo image
  super-resolution.
\newblock {\em arXiv preprint arXiv:2303.13807}, 2023.

\bibitem{huang2016deep}
Gao Huang, Yu Sun, Zhuang Liu, Daniel Sedra, and Kilian~Q Weinberger.
\newblock Deep networks with stochastic depth.
\newblock In {\em Computer Vision--ECCV 2016: 14th European Conference,
  Amsterdam, The Netherlands, October 11--14, 2016, Proceedings, Part IV 14},
  pages 646--661. Springer, 2016.

\bibitem{Jeon}
Daniel~S Jeon, Seung-Hwan Baek, Inchang Choi, and Min~H Kim.
\newblock Enhancing the spatial resolution of stereo images using a parallax
  prior.
\newblock In {\em Proceedings of the IEEE conference on computer vision and
  pattern recognition}, pages 1721--1730, 2018.

\bibitem{VDSR}
J. {Kim}, J.~K. {Lee}, and K.~M. {Lee}.
\newblock Accurate image super-resolution using very deep convolutional
  networks.
\newblock In {\em 2016 IEEE Conference on Computer Vision and Pattern
  Recognition (CVPR)}, pages 1646--1654, 2016.

\bibitem{DRCN}
J. {Kim}, J.~K. {Lee}, and K.~M. {Lee}.
\newblock Deeply-recursive convolutional network for image super-resolution.
\newblock In {\em 2016 IEEE Conference on Computer Vision and Pattern
  Recognition (CVPR)}, pages 1637--1645, 2016.

\bibitem{IMSSRNet}
Jianjun Lei, Zhe Zhang, Xiaoting Fan, Bolan Yang, Xinxin Li, Ying Chen, and
  Qingming Huang.
\newblock Deep stereoscopic image super-resolution via interaction module.
\newblock {\em IEEE Transactions on Circuits and Systems for Video Technology},
  31(8):3051--3061, 2020.

\bibitem{eformer}
Wenbo Li, Xin Lu, Jiangbo Lu, Xiangyu Zhang, and Jiaya Jia.
\newblock On efficient transformer and image pre-training for low-level vision.
\newblock {\em arXiv preprint arXiv:2112.10175}, 2021.

\bibitem{liang2021swinir}
Jingyun Liang, Jiezhang Cao, Guolei Sun, Kai Zhang, Luc Van~Gool, and Radu
  Timofte.
\newblock Swinir: Image restoration using swin transformer.
\newblock In {\em Proceedings of the IEEE/CVF International Conference on
  Computer Vision}, pages 1833--1844, 2021.

\bibitem{EDSR}
B. {Lim}, S. {Son}, H. {Kim}, S. {Nah}, and K.~M. {Lee}.
\newblock Enhanced deep residual networks for single image super-resolution.
\newblock In {\em 2017 IEEE Conference on Computer Vision and Pattern
  Recognition Workshops (CVPRW)}, pages 1132--1140, 2017.

\bibitem{lu2022transformer}
Zhisheng Lu, Juncheng Li, Hong Liu, Chaoyan Huang, Linlin Zhang, and Tieyong
  Zeng.
\newblock Transformer for single image super-resolution.
\newblock In {\em Proceedings of the IEEE/CVF Conference on Computer Vision and
  Pattern Recognition}, pages 457--466, 2022.

\bibitem{nonlocal}
Yiqun Mei, Yuchen Fan, and Yuqian Zhou.
\newblock Image super-resolution with non-local sparse attention.
\newblock In {\em Proceedings of the IEEE/CVF Conference on Computer Vision and
  Pattern Recognition}, pages 3517--3526, 2021.

\bibitem{CSNLN}
Yiqun Mei, Yuchen Fan, Yuqian Zhou, Lichao Huang, Thomas~S Huang, and Honghui
  Shi.
\newblock Image super-resolution with cross-scale non-local attention and
  exhaustive self-exemplars mining.
\newblock In {\em Proceedings of the IEEE/CVF conference on computer vision and
  pattern recognition}, pages 5690--5699, 2020.

\bibitem{KITTI2015}
Moritz Menze and Andreas Geiger.
\newblock Object scene flow for autonomous vehicles.
\newblock In {\em Proceedings of the IEEE conference on computer vision and
  pattern recognition}, pages 3061--3070, 2015.

\bibitem{HAN}
Ben Niu, Weilei Wen, Wenqi Ren, Xiangde Zhang, Lianping Yang, Shuzhen Wang,
  Kaihao Zhang, Xiaochun Cao, and Haifeng Shen.
\newblock Single image super-resolution via a holistic attention network.
\newblock In {\em Computer Vision--ECCV 2020: 16th European Conference,
  Glasgow, UK, August 23--28, 2020, Proceedings, Part XII 16}, pages 191--207.
  Springer, 2020.

\bibitem{Middlebury}
Daniel Scharstein, Heiko Hirschm{\"u}ller, York Kitajima, Greg Krathwohl, Nera
  Ne{\v{s}}i{\'c}, Xi Wang, and Porter Westling.
\newblock High-resolution stereo datasets with subpixel-accurate ground truth.
\newblock In {\em Pattern Recognition: 36th German Conference, GCPR 2014,
  M{\"u}nster, Germany, September 2-5, 2014, Proceedings 36}, pages 31--42.
  Springer, 2014.

\bibitem{song}
Wonil Song, Sungil Choi, Somi Jeong, and Kwanghoon Sohn.
\newblock Stereoscopic image super-resolution with stereo consistent feature.
\newblock In {\em Proceedings of the AAAI Conference on Artificial
  Intelligence}, volume~34, pages 12031--12038, 2020.

\bibitem{AiAyn}
Ashish Vaswani, Noam Shazeer, Niki Parmar, Jakob Uszkoreit, Llion Jones,
  Aidan~N Gomez, {\L}ukasz Kaiser, and Illia Polosukhin.
\newblock Attention is all you need.
\newblock {\em Advances in neural information processing systems}, 30, 2017.

\bibitem{ntire2022}
Longguang Wang, Yulan Guo, Yingqian Wang, Juncheng Li, Shuhang Gu, Radu
  Timofte, Liangyu Chen, Xiaojie Chu, Wenqing Yu, Kai Jin, et~al.
\newblock Ntire 2022 challenge on stereo image super-resolution: Methods and
  results.
\newblock In {\em Proceedings of the IEEE/CVF Conference on Computer Vision and
  Pattern Recognition}, pages 906--919, 2022.

\bibitem{PASSR}
Longguang Wang, Yulan Guo, Yingqian Wang, Zhengfa Liang, Zaiping Lin, Jungang
  Yang, and Wei An.
\newblock Parallax attention for unsupervised stereo correspondence learning.
\newblock {\em IEEE transactions on pattern analysis and machine intelligence},
  44(4):2108--2125, 2020.

\bibitem{PASSRNet}
Longguang Wang, Yingqian Wang, Zhengfa Liang, Zaiping Lin, Jungang Yang, Wei
  An, and Yulan Guo.
\newblock Learning parallax attention for stereo image super-resolution.
\newblock In {\em Proceedings of the IEEE/CVF Conference on Computer Vision and
  Pattern Recognition}, pages 12250--12259, 2019.

\bibitem{man}
Yan Wang, Yusen Li, Gang Wang, and Xiaoguang Liu.
\newblock Multi-scale attention network for single image super-resolution.
\newblock {\em arXiv preprint arXiv:2209.14145}, 2022.

\bibitem{Flickr1024}
Yingqian Wang, Longguang Wang, Jungang Yang, Wei An, and Yulan Guo.
\newblock Flickr1024: A large-scale dataset for stereo image super-resolution.
\newblock In {\em Proceedings of the IEEE/CVF International Conference on
  Computer Vision Workshops}, pages 0--0, 2019.

\bibitem{iPASSR}
Yingqian Wang, Xinyi Ying, Longguang Wang, Jungang Yang, Wei An, and Yulan Guo.
\newblock Symmetric parallax attention for stereo image super-resolution.
\newblock In {\em Proceedings of the IEEE/CVF Conference on Computer Vision and
  Pattern Recognition (CVPR) Workshops}, pages 766--775, June 2021.

\bibitem{uformer}
Zhendong Wang, Xiaodong Cun, Jianmin Bao, Wengang Zhou, Jianzhuang Liu, and
  Houqiang Li.
\newblock Uformer: A general u-shaped transformer for image restoration.
\newblock In {\em Proceedings of the IEEE/CVF Conference on Computer Vision and
  Pattern Recognition}, pages 17683--17693, 2022.

\bibitem{xu2021deep}
Qingyu Xu, Longguang Wang, Yingqian Wang, Weidong Sheng, and Xinpu Deng.
\newblock Deep bilateral learning for stereo image super-resolution.
\newblock {\em IEEE Signal Processing Letters}, 28:613--617, 2021.

\bibitem{yan2020disparity}
Bo Yan, Chenxi Ma, Bahetiyaer Bare, Weimin Tan, and Steven~CH Hoi.
\newblock Disparity-aware domain adaptation in stereo image restoration.
\newblock In {\em Proceedings of the IEEE/CVF Conference on Computer Vision and
  Pattern Recognition}, pages 13179--13187, 2020.

\bibitem{drfn}
Xin Yang, Haiyang Mei, Jiqing Zhang, Ke Xu, Baocai Yin, Qiang Zhang, and
  Xiaopeng Wei.
\newblock Drfn: Deep recurrent fusion network for single-image super-resolution
  with large factors.
\newblock {\em IEEE Transactions on Multimedia}, 21(2):328--337, 2018.

\bibitem{Ressam}
Xinyi Ying, Yingqian Wang, Longguang Wang, Weidong Sheng, Wei An, and Yulan
  Guo.
\newblock A stereo attention module for stereo image super-resolution.
\newblock {\em IEEE Signal Processing Letters}, 27:496--500, 2020.

\bibitem{swinfir}
Dafeng Zhang, Feiyu Huang, Shizhuo Liu, Xiaobing Wang, and Zhezhu Jin.
\newblock Swinfir: Revisiting the swinir with fast fourier convolution and
  improved training for image super-resolution.
\newblock {\em arXiv preprint arXiv:2208.11247}, 2022.

\bibitem{RCAN}
Yulun Zhang, Kunpeng Li, Kai Li, Lichen Wang, Bineng Zhong, and Yun Fu.
\newblock Image super-resolution using very deep residual channel attention
  networks.
\newblock In {\em Proceedings of the European conference on computer vision
  (ECCV)}, pages 286--301, 2018.

\bibitem{RDN}
Yulun Zhang, Yapeng Tian, Yu Kong, Bineng Zhong, and Yun Fu.
\newblock Residual dense network for image super-resolution.
\newblock In {\em Proceedings of the IEEE conference on computer vision and
  pattern recognition}, pages 2472--2481, 2018.

\end{thebibliography}
}

\end{document}